\pdfoutput=1

\documentclass[11pt]{article}

\usepackage[]{acl}

\usepackage{times}
\usepackage{latexsym}
\usepackage{graphics}

\usepackage{graphicx}

\usepackage[T1]{fontenc}

\usepackage[utf8]{inputenc}

\usepackage{microtype}

\title{Summarizing a virtual robot's past actions in natural language}

\author{Chad DeChant \\
  Columbia University \\
  \texttt{chad.dechant@columbia.edu} \\\And
  Daniel Bauer \\
  Columbia University\\
  \texttt{bauer@cs.columbia.edu} \\}

\begin{document}
\maketitle
\begin{abstract}
We propose and demonstrate the task of giving natural language summaries of the actions of a robotic agent in a virtual environment. We explain why such a task is important, what makes it difficult, and discuss how it might be addressed. To encourage others to work on this, we show how a popular existing dataset that matches robot actions with natural language descriptions designed for an instruction following task can be repurposed to serve as a training ground for robot action summarization work. We propose and test several methods of learning to generate such summaries, starting from either egocentric video frames of the robot taking actions or intermediate text representations of the actions used by an automatic planner. We provide quantitative and qualitative evaluations of our results, which can serve as a baseline for future work.   \end{abstract}

\section{Introduction}
As artificially intelligent agents, particularly robots, become more capable and are entrusted with more tasks, it will be increasingly important to reliably keep track of what they do. However, robots in the physical world as well as non-embodied agents will routinely perform roles that make direct supervision of them difficult or impossible. A robot may, for example, be used to move many loads of construction material from place to place, or a non-embodied agent may engage in hundreds of text-based dialogues with customers over the course of a day. In both cases, real time human oversight would be impractical. It will therefore be necessary to develop methods to monitor and record the actions of such agents and provide that recorded information at a later time to a human operator or user. One way to do that is to develop the capability for robots and other agents to report on and summarize their actions in natural language.

It might be thought that a robot or other agent could simply keep a log of all it has done and thus give a full report of every turn, move, and decision it made during its operation. However, there are two problems with such a scenario. First, such reporting may not be possible, particularly if an agent's actions are not the result of interpretable internal planning using discrete primitives but are instead the result of following a reinforcement learning policy implemented as a neural network which simply outputs, for example, rotations and joint movements to a robot's wheels, arms, etc. Second, even if such a complete record of action existed and was human readable, it would not be useful; it would be far too long and detailed to read and make sense of in a reasonable time in any realistic situation. Instead, it will be necessary for agents to summarize their activities. And in order to make such a summary available in a format for humans to comprehend quickly and accurately it would be ideal if the summary were given in natural language.

\begin{figure*}
\begin{center}
\centerline{\includegraphics[width=6in]{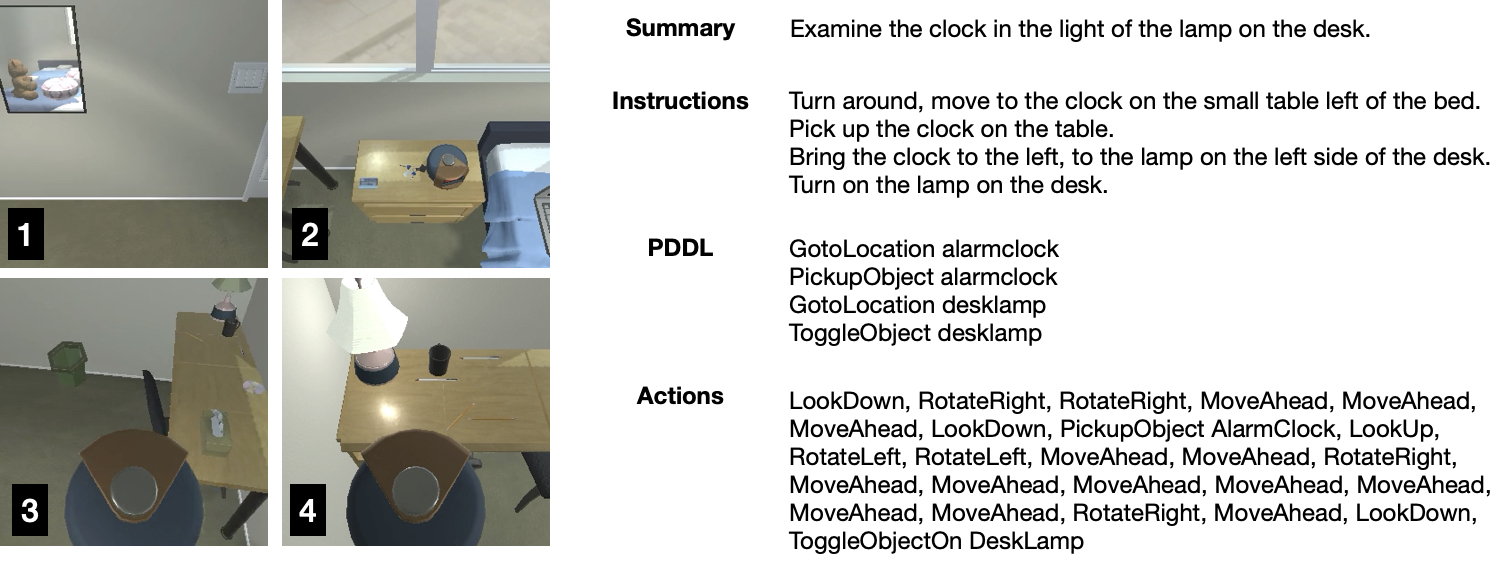}}
    \caption{Sample selection of video frames (numbered to indicate their order), high level summary, instructions, high level \textsc{pddl}, and our simplified version of the low level actions for one human annotation of one episode in the \textsc{alfred} dataset's training set.}
\label{alfreddatasetimg}
\end{center}
\end{figure*}

Summarizing past robotic actions is related to a thriving research area, natural language instruction following, to which it can be seen as a complement. A user who gave such instructions might naturally be expected to want a short natural language summary of what was done in response to those instructions. Yet despite the volume of work that has been done on instruction following, its complement has gone largely unaddressed. We present the first work directly addressing, performing, and evaluating robot action summarization.

To do so we take advantage of the complementary nature of action summarization and instruction following by repurposing the popular \textsc{alfred} dataset \citep{shridhar2020alfred} intended to train virtual robotic agents to perform complicated, multi-step actions following natural language instructions in English. The dataset contains egocentric videos of robotic agents performing actions along with one sentence summary descriptions of the actions and multi-sentence step-by-step descriptions intended to serve as instructions. The dataset also contains high and low level descriptions in near-natural language of the actions as worked out by an automatic planner when creating the dataset. Figure \ref{alfreddatasetimg} illustrates one episode in the dataset along with the various representations available.

Our experiments involve different combinations of the representation inputs available in the dataset, corresponding to the ways a robotic agent may have stored elements in memory. First, on the assumption that a robotic agent may be able to produce a record of its actions in the form of a discrete, text-based planning language as it performs them (such as \textsc{pddl}\footnote{Planning Domain Description Language}), we find that these can very successfully be transformed into an accurate natural language summary. Second, although our focus is on generating the short summaries, we also attempt to generate the more detailed descriptions intended to serve as instructions in the original \textsc{alfred} challenge, which we find are more difficult to accurately produce. We use a T5 encoder-decoder model for all text-to-text experiments. 
Third, if we cannot rely on the availability of discrete text-based descriptions of actions, we show that we are able to generate summaries using frames from the videos captured by a robotic agent in the course of its operation, though these summaries are not as accurate as those derived from the planning text. We use pre-trained \textsc{CNNs} to extract visual features and a bidirectional \textsc{RNN} decoder to produce the summary. We find that these summaries can be improved if we use a multi-step pipeline that takes in the video frames, first produces a text description of such actions in the intermediate planning language, and then transforms that text description into natural language. Finally, we also examine the multimodal case of generating summaries by using both text and video frames as input.

After surveying related work in section \ref{sec:related}, we discuss how we repurpose the \textsc{alfred} dataset in section \ref{sec:dataset}. In section \ref{sec:stages}, we discuss the task of robot action summarization in general and propose how to break it down into stages, discussing the challenges each stage presents. We then present our main experiments and results in section \ref{sec:experiments} and discuss our findings in section \ref{sec:discussion}. Section \ref{sec:conclusion} concludes the paper.

\section{Related Work}
\label{sec:related}
\noindent
\textbf{RoboNLP \& Natural language instruction \\following}
\noindent

\citet{dechant2021toward} proposed robot action summarization as a research direction, suggesting a set of tasks to pursue, but did not perform any experiments; this work could be considered to fall under the first of those tasks, pure summarization of past robotic action.

\citet{tangiuchi2019survey} and  \citet{tellex2020robots} offer thorough reviews of language use in the context of robotics. Some prior work has been done on using natural language to clarify aspects of how a simple action was performed in response to a question \citep{yoshino2021caption}. Detailed descriptions of actions such as robots playing soccer \citep{mooney2008learning} or automated driving \citep{barrett2015robot, barrett2017driving} have been generated. These have not involved learning how to report and condense a series of actions into anything like a summary, however. 

Our proposal is closely related to learning to follow natural language instructions, which has generated a great deal of interest at the intersection of robotics and natural language processing. \citet{winograd1972understanding} demonstrated an early system to provide instructions to and interact with a robotic agent in a virtual 'blocks world' environment using natural language. Recently, \citet{shridhar2021cliport} train a robotic arm in a virtual environment to perform a range of tasks following natural language instructions and transfer the learned model to a real world robot. \citet{calvin21} introduce a benchmark for so-called long horizon (many step) robotic manipulation tasks following natural language instructions.

Rich simulated environments for navigation tasks have been introduced in recent years. \citet{anderson2018vision} introduced the Room to Room vision and language navigation dataset, which became the basis for much work in this area. Some of that work has involved learning to generate natural language descriptions of navigation trajectories as a training signal or tool: \citet{nguyen-etal-2020-iterative} provide feedback to an agent in the Room to Room environment by describing in natural language the paths the agent actually takes so it can learn to compare that to the path it should have taken; \citet{fried2018speaker} learn to generate instructions to augment training data and then, at test time, to evaluate the similarity of routes it might take with the description of the route it is supposed to take.  \citet{ramakrishnan2021exploration} offer a survey of work on embodied agents exploring a virtual environment and learning to interpret visual data. 

The \textsc{alfred} dataset \citep{shridhar2020alfred} we repurpose and will describe in the next section in detail has inspired a great deal of work on its natural language instruction following challenge. \citet{shridhar2021alfworld} improve an agent's ability to perform tasks in the virtual environment by first training them to learn to act in the interactive text only TextWorld environment \citep{cote2018textworld} in similar situations which are there described there only in text. \citet{pashevich2021episodic} learn to leverage the presence of the high level \textsc{pddl} plans to produce better representations of the natural language instructions by also training those representations to be used to generate \textsc{pddl} plans from the natural language instructions. Note that this not something we propose to do here; indeed, it is exactly the inverse operation of one of our experiments presented in section 3.2.

\noindent
\textbf{Summarization}

\noindent
There is an extensive body of work on natural language summarization, providing examples and resources for the new but related task of robot action summarization (see \citet{nenkova2012survey} and \citet{gambhir2017recent} for reviews). There are two main kind of summarization. In extractive summarization, the summaries are selected from the original text already present in a source document. In abstractive summarization, by contrast, new text is generated as the summary, allowing for a higher level of description. We employ an abstractive approach here, using recurrent sequence to sequence models similar to those which have successfully been used for summarization \citep{rush2015neural}. We also follow recent work using the Transformer architecture \citep{vaswani2017attention} to perform abstractive summarization \citep{lewisbart, raffel2020exploring}.

Because we are primarily interested in very short summaries of action sequences, our work is in some ways similar to so-called extreme summarization \citep{narayan2018don}. Future extensions of our work will require processing longer or multiple successive action episodes so will benefit from work on summarizing long texts such as novels \citep{wu2021recursively} or multiple documents \citep{liu-lapata-2019-hierarchical}.

\noindent
\textbf{Grounding language}

It has been recognized for some time that grounding language to the real world is essential for creating \textsc{ai} systems that could actually understand the language they processed \citep{harnad1990symbol}. Recent proposals on the need to situate natural language processing in a grounded or embodied context have brought renewed attention to this issue \citep{bisk2020experience, chandu2021grounding, mcclelland2020placing, lake2021word}. Though these did not discuss robots summarizing their actions, our work is a contribution to this direction of research.

Work on understanding video is relevant to our work since we are interested in using video or selected images from video as one of the inputs to summarizing a robot's action in natural language. The task of 'video summarization' in the computer vision community refers to selecting important frames of a video that can, together, serve as a visual summary of the whole video; see \citet{apostolidis2021video} for a current review of such techniques. Some work has been done on multimodal summarization from video and text transcripts to natural language summaries; \citet{Palaskar2019MultimodalAS} is one example, going from video and text in the How2 video dataset \citep{Sanabria2018How2AL} to summaries. \citet{jangra2021survey} provides a thorough survey of ``multimodal summarization'' of all kinds.

\section{Repurposed dataset}
\label{sec:dataset}
Because the task of robot action summarization is a new one there is not an existing dataset explicitly designed for it. Fortunately, however, there is a popular dataset which we can repurpose for this task. The \textsc{alfred} dataset is used to train robotic agents embodied in the virtual AI-2 Thor \citep{kolve2017ai2} environment to perform complicated, multi-step tasks given natural language instructions.\footnote{The dataset is available online at \url{https://github.com/askforalfred/alfred} under an \textsc{mit} license.} Its associated challenge, evaluated on a held-out test set unavailable to the public, is quite difficult; the best performing model listed on the challenge's leaderboard as of January 2022 achieves only a 40\% success rate in environments the model was trained on and 34\% in unseen environments (which is a significant improvement from the dataset's creators' baseline model's respective 4\% and 0.39\%).

Figure \ref{alfreddatasetimg} illustrates one example of an episode in the dataset, along with the various types of task representation the dataset includes. These task representations are:

• \emph{Summaries}: Natural language one sentence summaries of the whole action sequence. These are referred to as \emph{goal descriptions} in the original dataset.

• \emph{Instructions}: Natural language step by step instructions over several sentences. These can be seen as more detailed descriptions of the agent's actions than the very high level single sentence summaries.

• \emph{High \textsc{pddl}}: High level action plan in the Planning Domain Description Language \citep{aeronautiques1998pddl} with semantically rich descriptions. 

• \emph{Action descriptions}: Low level action plan generated by an automatic planner corresponding to available actions in the environment. It is more detailed and longer than High \textsc{pddl} but less easily readable and contains slightly less semantically rich content. For example, where the higher level \textsc{pddl} might read "GotoLocation alarmclock" the lower level actions might be a sequence of "MoveAhead", "RotateRight", and "RotateLeft" actions. 

• \emph{Video, images, and visual features}: Raw video of a task episode as well as still frames from the video and extracted intermediate features of a selection of the still frames as processed by a Resnet-50 convolutional neural network \citep{he2016deep}

The \textsc{alfred} dataset and its accompanying benchmark were designed to train and test an agent's ability to follow natural language instructions to perform tasks. We propose inverting that order, instead using it to train an agent to summarize the actions it takes during the tasks. Since the natural language instructions and summaries (goal descriptions) were actually generated after the fact by humans watching videos of what the robotic agents did, these can naturally be treated as descriptions or summaries of the action sequences. 

The publicly available \textsc{alfred} dataset does not include the test set so we do not use the test set here. The dataset contains a training set and two validation sets. One of the validation sets contains episodes in virtual environments which are present in the training set (\emph{Valid Seen}) while the other presents episodes in only previously unseen environments (\emph{Valid Unseen}). Note that the task types themselves and the kinds of objects present (e.g. sinks, apples) are the same in both validation sets, though their appearances may be different. We treat the Valid Seen set as our validation set, choosing models that perform best on it without regard to their performance on the Valid Unseen set, which therefore behaves as our test set.

The actual episodes and the low level action plans that describe them in the training set are not present in either validation set but there is a good deal of overlap between examples of high level \textsc{pddl} in the training and validation sets. In all cases involving high level \textsc{pddl} we therefore remove the overlaps from the valid seen and valid unseen sets, ensuring that there are no duplicates between the train and validation sets. The smaller valid seen set contains 244 annotations while the valid unseen contains 298, compared with 820 and 821 in the respective complete validation sets.

\section{The stages of robot action summarization}
\label{sec:stages}
 The overall challenge of a robotic agent's summarizing its past actions can be broken down into several stages, which can broadly be thought of as: choosing what goes into an agent's memory, storing those elements in memory, accessing the relevant memories at a later time, and then summarizing those accessed memories. The experiments in this paper focuses on the last stage, generating summaries of stored memories or representations of prior action. Here we discuss each stage, including the ways each stage is realized in this work as well as as some options for how it might be done differently.
 
\subsection{Choosing elements to remember} 
In order to form memories to call on later, two decisions must be made. First, what types of things will be remembered? And, second, which items of those types will be chosen to be remembered? 
The question of what types of things will be remembered will be largely determined by the percepts and internal states of the agents which will be storing memories. A robot will likely have multimodal perceptual inputs of some kinds, e.g. vision, language, sound, and tactile feedback. In our case, the robot agent has first-person perspective visual input in a simulated environment. An agent will also have an internal state, which may or may not itself be human interpretable. Here we have access to interpretable planning records in the form of high level \textsc{pddl} and lower level action plans. These \textsc{pddl} and action records document what the agent intended to do; in an environment where an action like \emph{MoveAhead} always happens as planned, these would be completely reliable. In the real world, or in a virtual environment with some degree of stochasticity, such plans would still be valuable sources of information but could not be assumed to be correct. It is also possible that agent could learn to generate something like the \textsc{pddl} descriptions of what actions it actually took if feedback from the environment were sufficiently meaningful, in which case such generated action descriptions could also be available for storing.

After determining what kinds of things can be stored as memories it will be necessary to determine which exemplars of those things will actually be stored by a robotic agent during operation. The simplest, though perhaps not the best, choice would be store everything, if that were possible given the capacity of the memory available to the agent. In this paper we examine the results of using different combinations of the available data but we restrict ourselves to using all of the \textsc{pddl} and action plans where we use them. We do not use every video frame or the full video stream. Instead, for the image percepts we use every video frame in the pre-selected smaller version of the \textsc{alfred} dataset curated by the \textsc{alfred} challenge creators, in which there is at least one video frame for every low level action. 

\subsection{Forming and accessing memories}
The mechanisms for storing and accessing memories could be as simple as saving image and text files to disk or as complicated as learning how to store and retrieve memories in a differential neural framework. We opt for the former approach and do not explore more complicated memory structures here, though there is much work that could be done on that in the future.

\subsection{Summarizing accessed memories}
Once memories are retrieved, they must be summarized. If the representations that form the memory are already in text form, whether natural language or close to it (such as the \textsc{pddl} plans), they can be summarized using techniques familiar in the natural language processing literature. There are two main classes of summarization. In \emph{extractive} summarization, a selection of the original elements are chosen and themselves form the summary while in \emph{abstractive} summarization, text is newly generated to serve as the summary, possibly producing higher level summaries than the extractive approach. If the items to be recalled are stored in a form other than text, it will be necessary to either translate them into natural language before summarizing them or to simply directly perform a kind of abstractive summarization on them to directly produce the summary. We do not use extractive techniques here so consider our approach to be abstractive, though in addition to some traditional summarization that involves only text we also perform abstractive summarization / conversion from action descriptions in a near-natural language intermediate planning language and images to natural language summaries.

\begin{table*}
\centering
\begin{tabular}{lc|cccc}
\hline
\textbf{Summarization input} & \textbf{No errors} & \textbf{Action errors} & \textbf{Object errors} &  \textbf{Place errors}& \textbf{Extra errors} \\
\hline

\textsc{pddl} & 98 & 0 & 0 & 2 & 0 \\
Actions & 96 & 0 & 4 & 0 & 0 \\
\hline
Generated \textsc{pddl} & 54 & 10 & 8 & 38 & 4 \\
Generated actions & 46 & 8 & 28 & 48 & 4 \\
\hline
Images only & 38 & 4 & 26 & 22 & 4 \\
\hline

Images and \textsc{pddl} & 56 & 4 & 14 & 34 & 0 \\
Images and actions & 52 & 0 & 20 & 22 & 10 \\

\hline
\end{tabular}
\caption{\label{tab:error-table}
Manual error analysis of high level summaries by method of generation. 
The percentage of generated examples with no errors is shown in the left column; in the right columns are percentages of errors present by type of error. Some examples may have more than one error.
}
\end{table*}

\section{Experiments and results}
\label{sec:experiments}
We present details of our experiments and their results broken down into sections devoted to the different combinations of possible elements to be remembered by a robot agent. These stored elements serve as the inputs to our various summarization approaches. We provide both automatic and manual evaluation metrics. 

First, we manually inspect a randomly selected set of fifty  examples for each summary output and calculate what percentage of these are error free or have one or more errors. We break down and report the possible errors into four types: \emph{action errors}, in which the main action that took place during the relevant episode is reported incorrectly in the generated summary; \emph{object errors}, in which the main object interacted with is wrongly named or counted; \emph{place errors}, in which the place or places mentioned are incorrect; and \emph{extra errors}, in which details which are unwarranted by the input are hallucinated by the model. These analyses can be found in Table ~\ref{tab:error-table}.

Second, we automatically calculate \textsc{rouge} \citep{lin-2004-rouge} and \textsc{bleu} \citep{papineni-etal-2002-bleu} scores for both the summaries and recovered instructions. These automatic evaluation scores can be found in Table ~\ref{tab:scores}. We report \textsc{rouge-1} (recall), \textsc{rouge-2} (recall), \textsc{rougle-l f1}, \textsc{bleu} and \textsc{bleu-1} scores.

\subsection{Text to text}
For all experiments where the input and the output are purely text, we fine tune a T5 pretrained large language model. The T5 language model, of which we use the "base" size configuration, was originally trained on several tasks, one of which was summarization \citep{raffel2020exploring}. We use a version of it available in the Hugging Face transformers library ('t5-base') \citep{wolf2020transformers}. \footnote{Our code will be available on Github.}

\noindent
\textbf{\textsc{pddl} or actions to summary}

If structured, intermediate representations such as the high level \textsc{pddl} and low level action plans were reliably available to record a robot's actions, they could be used as the basis of generating high level summaries, which is what we demonstrate here. The T5 model does a very good job of generating short summaries based on either high level \textsc{pddl} or low level actions as input. It makes an error on only one of our fifty manually inspected \textsc{pddl} to summary examples and two in the low level actions to summary examples. See Table ~\ref{tab:error-table} for full details.

\noindent
\textbf{\textsc{pddl} or actions to instructions}

We also use the T5 model to take as input either \textsc{pddl} or low level actions and attempt to recover the instructions, which provide more detailed step by step narration of each episode. This is a much more difficult task, as shown in the significantly lower performance numbers in Table ~\ref{tab:scores}. There is simply much more information in the natural language instructions than in either the high level \textsc{pddl} text or low level actions so both of the transformer models resort to inventing additional details to fill in the gaps. Upon visual inspection, these invented details are mostly reasonable guesses based on what the models have seen of the \textsc{alfred} environment but they are effectively guesses and are often wrong.

\begin{table*}
\centering
\begin{tabular}{lccccc|cccccc}
\hline
\textbf{Task} & & & \textbf{Seen} & & &  & &  \textbf{Unseen} & & \\
\hline

 & \textbf{R-1} & \textbf{R-2} &  \textbf{R-L}& \textbf{Bleu}  &
\textbf{Bleu-1} & \textbf{R-1} & \textbf{R-2} &  \textbf{R-L}& \textbf{Bleu} & \textbf{Bleu-1}\\
\hline

\textsc{pddl} to sum & .628 & .372 & .590 & .624 & .902 & .610 & .358 & .587 & .607 & .890\\
Actions to sum & .610 & .358 & .589 & .604 & .881 & .630 & .377 & .599 & .647 & .900\\
Gen \textsc{pddl} to sum & .596 & .344 & .565 & .580 & .877 & .518 & .271 & .505 & .472 & .810\\
Gen actions to sum & .555 & .301 & .537 & .485 & .826 & .514 & .269 & .491 & .425 & .760\\
\hline

\textsc{pddl} to inst & .557 & .325 & .529 & .529 & .866 & .545 & .310 & .519 & .527 & .869\\
Actions to inst & .566 & .329 & .528 & .539 & .854 & .570 & .338 & .527 & .551 & .867\\
Gen \textsc{pddl} to inst & .542 & .312 & .514 & .497 & .864 & .490	& .260	& .457 & .427 & .827\\
Gen actions to int & .508 & .279 & .488 & .462 & .844 & .493 & .270 & .457 & .433 & .826\\
\hline
Images to sum & .582 & .321 & .556 & .550 & .862 & .519 & .265 & .496 & .438 & .779\\
Images to inst & .540 & .314 & .496 & .501 & .805 & .536 & .292 & .460 & .438 & .769\\
Images to \textsc{pddl} & .923 & .881 & .923 & .854 & .942 & .761 & .597 & .763 & .594 & .824\\
Images to actions & .858 & .713 & .821 & .652 & .856 & .822 & .654 & .769 & .590 & .812\\
\hline
Img \& \textsc{pddl} to sum & .606 & .355 & .575 & .587 & .883 & .571 & .325 & .543 & .527 & .840\\
Img \& actions to sum & .572 & .321 & .549 & .524 & .843 & .519 & .269 & .498 & .417 & .785\\
Img \& \textsc{pddl} to inst & .563 & .329 & .498 & .514 & .830 & .542 & .286 & .451 & .437 & .792\\
Img \& actions to inst & .554 & .322 & .491 & .501 & .815 & .539 & .289 & .461 & .434 & .767\\
\hline
\end{tabular}
\caption{\label{tab:scores}
\textsc{rouge} and \textsc{bleu} scores for generated text. Results from virtual environments seen during training are on the left; unseen environments are on the right. (Img = Image; summ = summary; instr = instructions) 
}
\end{table*}

\subsection{Video frames to text}
To study going from visual input to text descriptions, we use the image features extracted from a pretrained Resnet model given selected video frames of an episode. These features, provided in the \textsc{alfred} dataset, consist of 512 $7 \times 7$ layers of features in the convolutional network. We pass these through two further convolutional layers and then into a bidirectional recurrent encoder-decoder network with attention, trained from scratch, to output text. Full details of this and the multimodal network we create can be found in Appendix A.

We experimented with two modifications to our visual pipeline, neither of which, surprisingly, had any significant effect. First, we changed the visual representations our network received by feeding the \textsc{jpeg} images of the dataset's video frames through a different pretrained convolutional neural network, the \textsc{cnn} used to encode images in the \textsc{clip} network \citep{radford2021learning}. Second, following the example of \citet{lu2021pretrained}, who showed that a pretrained language model could be used for some tasks in other modalities, we modified our convolutional neural networks to produce vector representations appropriately sized to feed into a T5 transformer network. In both cases, performance on going from video frames to generated \textsc{pddl} was not meaningfully different than the performance of the \textsc{rnn-cnn} network using Resnet features according to \textsc{rouge} and \textsc{bleu} scores.

\noindent
\textbf{Video frames to \textsc{pddl} and actions} 

We train our model to generate \textsc{pddl} and action descriptions using the image features as input. The model does a very good job on this task, as can be seen in Table ~\ref{tab:scores}. Evidence for the quality of the \textsc{pddl} and action description outputs also comes from the pipeline task described below which uses these as inputs to generate natural language descriptions. It might be noted that the lower level action plans contain a good number of 'MoveAhead' commands in each episode, which may have the effect of slightly flattering the results.

\noindent
\textbf{Video frames to summaries and instructions} 

Going directly from video frames to natural language summaries or instructions is less successful, as can be seen in the lower scores for both tasks in Table ~\ref{tab:scores}. It is also evident from the higher number of errors flagged by the manual inspection of the generated summaries in Table ~\ref{tab:error-table}. Generating summaries from video frames alone has by far the lowest rate of success; only 38\% of inspected generated summaries contain no errors at all.

\subsection{Pipeline of images to \textsc{pddl} or actions to text}

Because the \textsc{pddl} generated from video frames is, as judged by the automatic metrics, of significantly higher quality than the natural language summaries or instructions generated from images, we developed a pipeline that first takes in image features and produces the corresponding \textsc{pddl} or low level actions. These generated \textsc{pddl} or actions are then fed into the transformer models which we previously fine tuned to produce summaries or instructions from \textsc{pddl} or actions, respectively. We find that according to both automatic metrics and manual inspection of summaries the generated \textsc{pddl} and actions produce better output than is seen when going from video frames directly to summaries or instructions.\\

\noindent
\subsection{Multimodal image and text to text}

So far we have discussed generating summaries and instructions from either visual or text input. Finally, we combine both image features and high level \textsc{pddl} or low level actions to produce the natural language descriptions. To do this we modify our recurrent sequence to sequence architecture to accept both modalities at the same time by using a second bidirectional recurrent network to encode the text input and concatenating the resulting vector representation with the image representations before sending them to the decoder. Therefore, because the Transformer architectures we use for pure text to text tasks is not used, the appropriate comparison is to our purely vision-based convolutional recurrent network. We see that results are significantly better than the vision only method though fall short of the text to text Transformer models.

\section{Discussion}
\label{sec:discussion}
We have demonstrated that it is possible to develop the ability to summarize the past actions of a robotic agent operating in a virtual environment starting from various inputs. We found that a Transformer model could very reliably generate a natural language summary from high level \textsc{pddl} or low level action plans.  We also found that the vision model was reasonably good at generating such \textsc{pddl} or action descriptions, but did not perform as well when generating natural language summaries directly (end to end). A pipeline approach, using the text to text Transformer model to generate summaries from the \textsc{pddl} or action descriptions produced by a vision model, outperformed the end to end approach. This suggests that building upon the pipeline approach is a promising direction for future research. On the other hand, the results highlight the limitations of the vision component of our system. Since replacing the Resnet features with \textsc{clip cnn} features did not change the results, this suggests that a more significant change might be useful. The vision difficulties could to some extent be the result of the virtual environment our dataset makes use of.


While we have good results on generating the short one sentence summaries, which was the focus of our work here, the generated longer descriptions in the form of instructions were less reliably accurate. Further work will have to be done to allow for the generation of such more detailed descriptions, perhaps making more use of multimodal information when available.

In general we found that the automatic text generation quality metrics we used were aligned with our manual assessment upon inspection of the results. When significant errors were found in generated summaries they were most often because a word that should not have been there was, rather than that a word was missing. Missing words were often details which the human annotators of the \textsc{alfred} dataset included but which were relatively unimportant. This suggests that precision-oriented metrics such as \textsc{bleu} might be especially useful.

\noindent \textbf{Limitations}\\
The tasks which the virtual robotic agents perform in the \textsc{alfred} environment, while very challenging for the robotic agents to learn to perform, are relatively straightforward to describe in language; they do not present a great deal of variation, nor do they use a particularly wide range of vocabulary. Learning to summarize more diverse actions with a more diverse set of environments and objects would be another useful next step.

The sequences of actions undertaken in the dataset we use are considered quite long relative to other robotic tasks. However, they are quite short compared to what an actual robot operating the world would be expected to do. This means that our approach of using one image frame per low level action may not be realistic under certain approaches to the robot action summarization problem, necessitating further work on memory and selecting items for memory.

\section{Conclusion}
\label{sec:conclusion}
This work begins a line of research on robot action summarization. If robotic agents are to operate in the real world, as they already are, it is important that they be well supervised by humans and that their actions be understandable.
We suggest that establishing a basic narrative of \emph{what} an agent does is in some ways a prerequisite to understanding \emph{why} it does something. 



In future work, we plan to address the other stages of robot action summarization outlined in section \ref{sec:stages}, including choosing what to remember and how to represent it, and will extend our work to other datasets and robotic tasks. 
We will also investigate scenarios in which, rather than compiling a single static summary, a robotic agent needs to answer natural language questions about its past actions. Future work will also extend beyond reporting \emph{what} happened to explaining \emph{why} an agent did or failed to do something. 
We will also study how learning to summarize can be beneficial for learning to act. 

\subsection{Ethical considerations}

\noindent
Ethical considerations figure prominently in the identified need motivating the creation of this work. If we are to ensure that robots and other agents are operating safely and according to appropriate legal and ethical standards, it will be necessary at a minimum to understand and reliably track what they do. It is hoped and intended that this work will contribute to that supervision. 

Nevertheless, there may be some unintentional risks associated with further developments of this work. If users come to expect that a robot which provides a summary of its actions always does so accurately, they may not be prepared for errors or intentional deceit by, for example, someone who has hacked into the robot. So while it will be important to release summarization capabilities in real robots only after further work has been done to ensure accuracy, users will also have to be alerted to the fact that errors in reporting actions may still occur. It is also imaginable that a robot which could summarize its actions to its users could also summarize those actions to others, allowing for surveillance of it users and an invasion of privacy. Steps will have to be taken to protect a user's privacy.

\section*{Acknowledgements}
Thanks to Iretiayo Akinola, Zhou Yu, and Shuran Song for helpful discussion and feedback and to the Centre for Effective Altruism's Long-Term Future Fund for grant support.

\bibliography{anthology,custom}
\bibliographystyle{acl_natbib}

\appendix

\section{Neural network models and training details}
\label{sec:appendixa}

\subsection{Image to text CNN + RNN}

\begin{figure}[h]
\begin{center}
\centerline{\includegraphics[width=3in]{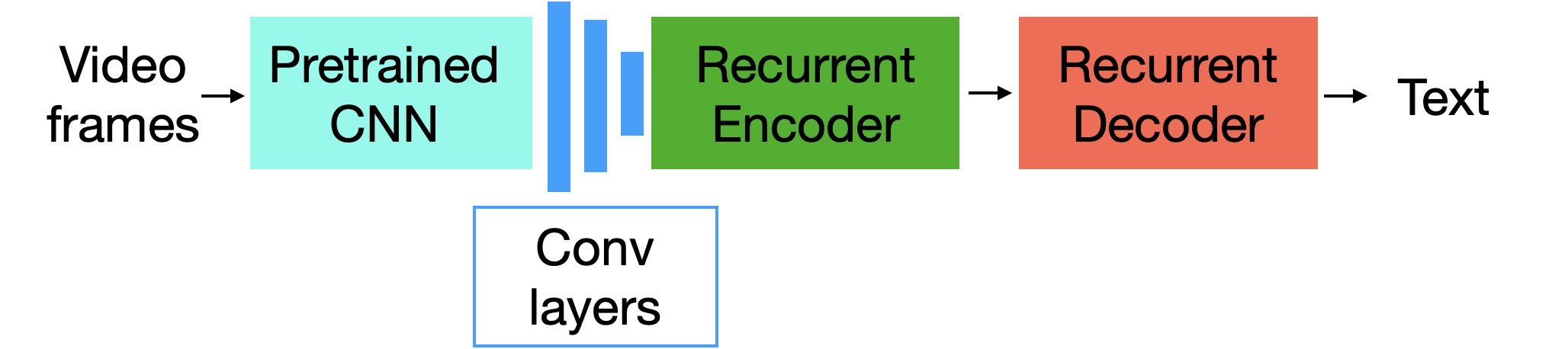}}
    \caption{Video frame to text CNN + RNN pipeline}
\label{pipeline1}
\end{center}
\end{figure}

\textbf{Encoder} 

Input: 512 x 7 x 7 Resnet features

Conv layer 1 (512, 128, 1)

Conv layer 2 (128, 32, 1)

Bidirectional GRU (3 layers, hidden size=512)

\textbf{Decoder}

Linear Layer (1536, 512)

GRU (hidden size = 512)

Attention over encoder outputs

Linear layer (1024, task dependent vocab size)

Number of parameters: 26,467,579

\subsection{Multimodal Image and text to text CNN/RNN + RNN}

\begin{figure}[h]
\begin{center}
\centerline{\includegraphics[width=3in]{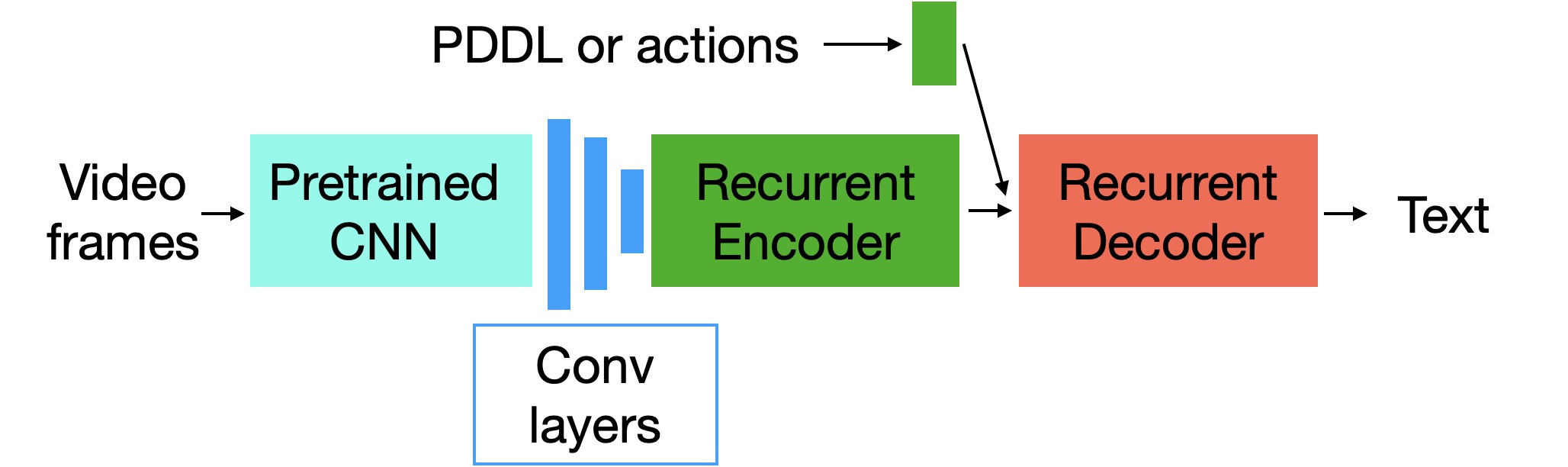}}
    \caption{Video frame and text (\textsc{pddl} or action) to text CNN + RNN pipeline}
\label{multimodalpipeline}
\end{center}
\end{figure}

\textbf{Image Encoder} 

Input: 512 x 7 x 7 Resnet features

Conv layer 1 (512, 128, 1)

Conv layer 2 (128, 32, 1)

Bidirectional GRU (3 layers, hidden size=512)

\textbf{Text Encoder} 
Bidirectional GRU (3 layers, hidden size=512)

\textbf{Decoder}

Linear Layer (3072, 512)

GRU (hidden size = 512)

Attention over image encoder outputs

Linear layer (1024, task dependent vocab size)

Number of parameters: 43,133,948

\subsection{Training information}

We primarily used \textsc{nvidia} Titan X \textsc{gpus}, up to three at a time. Though we did not track it, a rough estimate of total \textsc{gpu} time during initial exploration of this problem and the work reported here is 3000 hours.

Hyperparmeters we tested variations of include the optimizer (Adam, AdamW, AdaFactor, AdamW); the learning rate; network architecture choices (layer sizes, number of layers, batchnorm, dropout). The random seed was not one of the hyperparameters tuned.

As reported in the paper, we use the dataset's valid seen set as our validation set with which to choose hyperparameters. After choosing the hyperparameters we report the results of one run.

Our automatic metrics were generated from \textsc{blue} and \textsc{rouge} implementations available through the Hugging face library.

\end{document}